\documentclass[
]{ceurart}

\usepackage{listings}
\usepackage{latexsym}
\usepackage{amssymb}
\usepackage{amsmath}
\usepackage{amsthm}
\usepackage{booktabs}
\usepackage{enumitem}
\usepackage{graphicx}
\usepackage{color}
\usepackage{moresize}
\usepackage[normalem]{ulem}
\usepackage{soul} 
\usepackage{xcolor} 
\usepackage{tikz} 
\usepackage{pgfplots}
\usepackage{pgfplotstable}
\usepackage{orcidlink}
\usepackage{adjustbox}
\pgfplotsset{compat=1.8}

\newcommand{\BibTeX}{B\kern-.05em{\sc i\kern-.025em b}\kern-.08em\TeX}

\begin{document}

\copyrightyear{2024}
\copyrightclause{Copyright for this paper by its authors.
  Use permitted under Creative Commons License Attribution 4.0
  International (CC BY 4.0).}

\conference{MAI-XAI'24: Multimodal, Affective and Interactive eXplainable AI,
  October 19--20, 2024, Santiago de Compostela, Spain}

\title{From Feature Importance to Natural Language Explanations Using LLMs with RAG}


\author[1]{Sule Tekkesinoglu}[%
  orcid=0000-0002-1232-346X,
  email=sule@robots.ox.ac.uk,
  ]
\cormark[1]
\address[1]{Oxford Robotics Institute, University of Oxford, Oxford, UK}

\author[1,2]{Lars Kunze}[%
  orcid=0000-0001-5302-1938,
  email=lars@robots.ox.ac.uk,
]
\address[2]{Bristol Robotics Laboratory, University of the West of England, Bristol, UK}

\cortext[1]{Corresponding author.}

\begin{abstract}
  As machine learning becomes increasingly integral to autonomous decision-making processes involving human interaction, the necessity of comprehending the model's outputs through conversational means increases. Most recently, foundation models are being explored for their potential as \textit{post hoc} explainers, providing a pathway to elucidate the decision-making mechanisms of predictive models. In this work, we introduce traceable question-answering, leveraging an external knowledge repository to inform the responses of Large Language Models (LLMs) to user queries within a scene understanding task. This knowledge repository comprises contextual details regarding the model's output, containing high-level features, feature importance, and alternative probabilities. We employ subtractive counterfactual reasoning to compute feature importance, a method that entails analysing output variations resulting from decomposing semantic features. Furthermore, to maintain a seamless conversational flow, we integrate four key characteristics -- social, causal, selective, and contrastive -- drawn from social science research on human explanations into a single-shot prompt, guiding the response generation process. Our evaluation demonstrates that explanations generated by the LLMs encompassed these elements, indicating its potential to bridge the gap between complex model outputs and natural language expressions.
\end{abstract}



\maketitle

\section{Introduction}

As we become increasingly reliant on AI applications in our daily lives, it becomes imperative to interact with autonomous decision-making systems in human-understandable terms~\cite{arrieta2020explainable,kaur2022trustworthy}. Regulatory frameworks and standards are also evolving to require transparency and accountability in AI-driven systems~\cite{varovsanec2022path,nannini2023explainability}. The technical necessity for explanations to facilitate model debugging and rectifying potential ethical and legal risks arising from biases and errors has been extensively discussed in the literature~\cite{doshi2017towards,hassija2024interpreting,vale2022explainable}.

The increasing demand for transparent and interpretable AI models has prompted the exploration of various explanation methods, resulting in significant progress in both inherently interpretable and post-hoc explainability methods~\cite{slack2021reliable,madsen2022post,linardatos2020explainable}. While effective, these methods struggle to provide explicit interpretability or intuitive explanations for non-technical users. Consequently, there has been a concerted effort to integrate interpretability approaches with Natural Language Processing (NLP)~\cite{cambria2023survey,chen2023driving}. Recently, Large Language Models (LLMs) have gained acclaim as post-hoc explainers, showcasing their potential to elucidate decisions made by other predictive models. Despite their success, a persistent challenge for these models is the occurrence of object hallucinations—instances where plausible yet incorrect outputs are generated, such as featuring objects that do not exist in the images~\cite{cui2024survey}. This raises questions about the fidelity of explanations to the underlying model. To address this issue, we propose a traceable question-answering, which informs LLM responses through an external knowledge source that provides insights into the model's output. By utilizing this data, the LLM acts as a reasoning engine to process the information rather than a source of information. Figure 1 provides a high-level depiction of this process.  

Additionally, to cultivate a seamless conversational experience, we incorporated key elements derived from social science research on human explanations into the system prompt, directing the response generation process. We present our vision for creating this integrated approach, outlining each component, and illustrating its viability through a real-world dataset. This integration has the potential to bridge the gap between intricate model outputs and user comprehension towards advancing the development of human-understandable explanations.

\begin{figure}[htb!]

\centering
\includegraphics[width=0.8\textwidth]{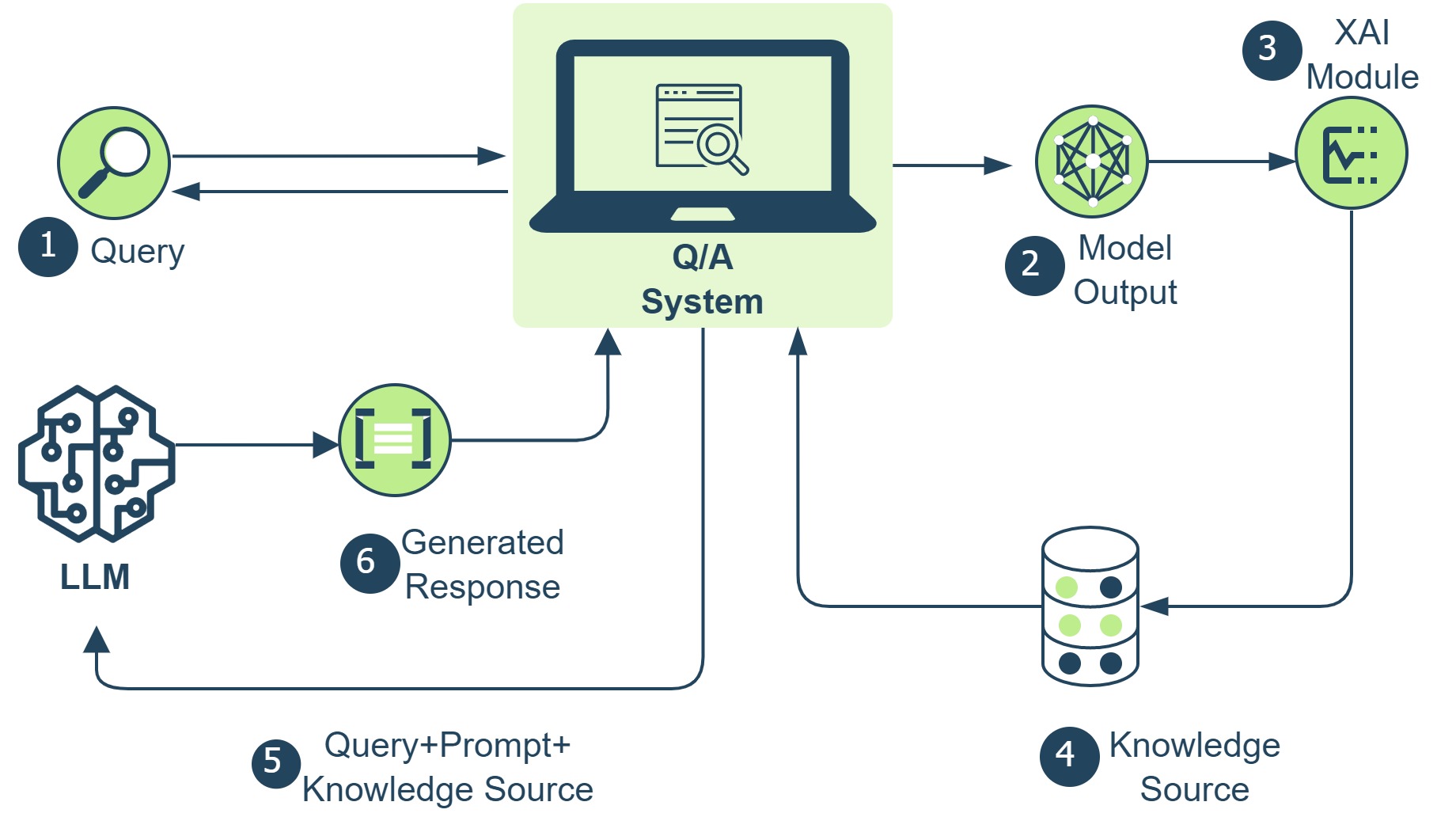}
\caption{Integration of post-hoc explainability approach with LLMs for traceable question-answering. The code is available at \url{https://github.com/suletekkesinoglu/XAI_LLM_RAG}}
\label{fig:rag}
\end{figure}

\section{Background}
\subsection{Recent advances in image explanations}
The demand for transparent and interpretable AI models has led to exploring diverse explanation methods for vision tasks \cite{zeiler2014visualizing}. Recent years have witnessed significant progress in both intrinsic and post-hoc explainability methods. Gradient-based techniques, which analyse the gradients of the model's output with respect to its input features, have been extensively studied \cite{selvaraju2017grad}. While effective, these methods may fall short in providing intuitive explanations for non-experts. In another line of work, attention mechanisms are proposed by assigning weights to different parts of the input data based on their relevance to the model's decision \cite{bahdanau2014neural}. Similarly, attention mechanisms may lack explicit interpretability, and some argue that attention alone is insufficient. As a result, there have been concerted efforts to integrate these approaches with Natural Language Processing (NLP) \cite{chen2023driving}. Long Short-Term Memory (LSTM) encoders have been employed to extract features from video clips, with LSTM decoders generating descriptive texts based on these features. While effective in image captioning, these methods provide textual descriptions rather than visual explanations. Moreover, researchers have proposed perturbation-based approaches—a model-agnostic method that perturbs input data and observes changes in model predictions. Building on this research, we integrate semantic perturbation analysis with pretrained language models, translating the results into the text modality for enhanced explainability.  

\subsection{LLMs as post hoc explainers}
Recent strides in Large Language Models (LLMs), exemplified by GPT-4 (OpenAI), Bard (Google), Claude-2 (Anthropic), and Llama-2 (Meta), mark a transformative era in Natural Language Processing (NLP) research. These models have become pervasive across diverse applications, ranging from machine translation and question-answering to text generation~\cite{hadi2023survey}. Notably, LLMs have recently gained attention as post hoc explainers, highlighting their potential to elucidate the decisions made by other predictive models~\cite{kroeger2023large}. Leveraging their in-context learning capabilities by fine-tuning for specific tasks enhances their proficiency in generating contextually relevant explanations \cite{chen2023driving,yang2024fine}. Despite their success, a persistent challenge for these models is the occurrence of object hallucinations—instances where plausible yet incorrect outputs are generated, featuring objects that do not exist in the images in vision-based tasks. This issue raises questions about the faithfulness of explanations to the underlying model. One way to address this is by recognizing the role of the semantic map layer in capturing the meaning and context of the physical surroundings \cite{cui2024survey}. By effectively encoding valuable semantic information into LLMs, we can truthfully represent and comprehend the intricate details of the model output within the language space.

\subsection{Characteristics of explanations and XAI}
\label{sec:char}
In the context of XAI, the audience for explanations is predominantly human; thus, understanding what makes an explanation human-friendly is paramount. While accuracy is important, the clarity and presentation of explanations are equally crucial in ensuring comprehensibility. According to Miller~\cite{miller2019explanation}, current work on interpretability in machine learning relies solely on researchers' intuition regarding what constitutes an appropriate explanation for humans. Miller's survey highlights four major characteristics of explanations drawn from findings in social science research on human explanation: explanations are social, causal, selective, and contrastive~\cite{miller2019explanation}. In the following section, we briefly discuss these characteristics in relation to XAI before demonstrating how we integrate them into our application.

\subsubsection*{Explanations are social}
An explanation represents an interactive exchange between two roles: the explainer and the explainee, and is governed by certain `rules'~\cite{hilton1990conversational}. Interactive explanations in a dialogue structure consist of three fundamental components: opening statements, clarification questions, and closing statements~\cite{cawsey1993planning}. As such, it should follow the basic rules of conversation, which are captured by Grice's conversational maxims~\cite{grice1975logic}. These include principles such as quality, quantity, relation, and manner, which can be understood as `only say what you believe', `only say as much as is necessary', `only say what is relevant', and `say it nicely'. Manner further encompasses various sub-maxims, including avoiding unclear expression, eliminating ambiguity, maintaining brevity to avoid unnecessary wordiness, and presenting information in an orderly fashion. 

Moreover, explanatory agents employing anthropomorphic traits like politeness markers (e.g., ``Thank you!''), warm and friendly tones, and empathetic expressions (e.g., ``I understand.'') are pivotal for effective human interaction~\cite{feine2019taxonomy}. \textit{Theory of Mind} also plays a crucial role in social explanations.\footnote{Theory of mind refers to our ability to understand other people by attributing mental states to them, including beliefs, desires, intentions, emotions, and thoughts that may differ from our own~\cite{apperly2009humans}.} As an integral part of a dialogue, an explanatory agent should keep track of what has already been explained, maintaining a simple mental model of others. This enables intelligent agents to tailor explanations based on past interactions and evolving contexts.

\subsubsection*{Explanations are causal}
Causal explanations aim to elucidate the cause-and-effect relationship underlying a decision. When explaining a prediction made by a machine learning model, identifying the most influential features can provide insight into the key causal connections to the prediction. This process, also known as causal inference,  starts with observations (e.g., what if a feature had a different value) and selecting some of those causes as the explanation. Miller divides casual inference into two parts: counterfactuals and abductive reasoning~\cite{miller2019explanation}.

The reasoner uses a model of hypothetical counterfactual cases to derive an explanation, such as perturbing inputs, to see how they causally impact the prediction (See Section~\ref{sec:subs_cont}). Abductive reasoning is inferring the causes from those observations to form the most probable explanation. One can think of abductive reasoning as the following process: observing unexpected or surprising events, generating one or more hypotheses about these events, judging the plausibility of the hypotheses, and selecting the `best' hypothesis as the explanation~\cite{hoffman2017explaining}. It is important to note that new information (e.g., different feature perturbations) can alter the most likely explanation.

\subsubsection*{Explanations are selected}

Explanations are contextual by nature. While an event may have numerous causes, the recipient of the explanation typically cares about only a specific subset relevant to the context~\cite{carvalho2019machine}. The explainer selects a subset based on various criteria called casual selection. Hilton~\cite{hilton2017social} argues that explanation selection is used for cognitive reasons since causal chains are often too large to comprehend. The criteria people use for explanation selection include abnormality, relevancy, and simplicity. 

Abnormality is a key criterion for explanation selection; as such, people select unusual causes to explain events~\cite{kahneman1981simulation}. In machine learning interpretability, identifying abnormal events corresponds to deviant or higher feature importance values. As pointed out within the conversational maxims, relevancy also plays a role in explanation selection---only say what is necessary and relevant. People select explanations to adhere to these maxims whilst persuading the explainee to new information or viewpoints. Another strong criterion in explanation selection is simplicity, which refers to those that cite fewer causes. People prefer uniquely sufficient causes that bring about the effect without any other causes, but there might be cases with multiple sufficient causes.

\subsubsection*{Explanations are contrastive}

Studies have shown that people only request contrastive explanations, essentially `why' questions framed to imply the differences between two possible outcomes~\cite{lipton1990contrastive}. As a result, explanations are typically offered in relation to the cause of one event compared to another that did not occur.

Similarly, in the context of ML interpretability, people are not specifically interested in all the factors that led to the prediction but instead in the factors that need to change so that the prediction would also change. Explanations that present some contrast between the instance to explain and a reference point (i.e., a hypothetical instance) are preferable because the cognitive burden of complete explanations is too high. However, most existing work considers contrastive questions but not contrastive explanations, providing two complete explanations individually. This could be because the contrastiveness is application-dependent, given that a reference point may not be apparent in all tasks~\cite{molnar2020interpretable}. For instance, in sentiment analysis or anomaly detection, contrastive cases might not be as straightforward as other tasks, such as banking, e.g., loan accepted vs rejected cases.

\section{Materials and Method}

This section outlines the key elements of the traceable question-answering methodology. We detail the proposed explainability approach for extracting feature importance and the prompting technique to generate accurate and human-friendly explanations. Additionally, we provide an application example illustrating how we establish a connection between the model output, the explainability technique, and language models to produce natural language explanations in an interactive format.


\subsection{Explanations through subtractive counterfactual reasoning}
\label{sec:subs_cont}

Subtractive counterfactual reasoning is the process of removing an event to understand its causal influence on an outcome~\cite{epstude2008functional}. Such reasoning is widely used as an XAI technique, exemplified by occlusion analysis and feature importance attribution, built on the same foundation - quantifying the variation in model output through systematic modifications of input components~\cite{anjomshoae2021context,zeiler2014visualizing,ivanovs2021perturbation,robnik2018perturbation}. There are distinctions in terms of how these methods remove features, elucidate model behaviour, and summarize feature importance. Nonetheless, they provide valuable insights into the relative significance of each feature influencing the decision-making process \cite{covert2021explaining,omeiza2022spoken,anjomshoae2021visual}. 

This study proposes a decomposition-based approach to examine the output variations by decomposing input values. By measuring such variations, we determine the degree of impact on an outcome and reason about the importance of each feature value. As a function, the model assigns a class label and prediction probability $f:x\rightarrow f(x)$ given input $y$. To measure the effect of each input value, we observe the model's prediction $p$ for $y$ without the knowledge of event $T_i=T_k$, where $T_k$ is the value of feature $T_i$. To explicitly represent the absence of information, the feature value $T_k$ is replaced with an undefined value, i.e., NaN (Not a Number). NaN values are treated as invalid or masked elements and are not considered in the computation. We attain an array containing probabilities for all perturbations resulting from decomposition. In order to determine the importance of an individual feature $T_i$, we begin by identifying the $max$ and $min$ values within this array, which establishes the decision boundaries for the given input. Next, we compute the feature importance as the position of the probability-- after decomposing the feature-- relative to the $max$ and $min$ probabilities. The computation is expressed as follows:

\begin{equation}
importance(T_i) = \frac{p(x|y \setminus T_k)_{max} - p(x|y \setminus T_k)}{p(x|y \setminus T_k)_{max} - p(x|y \setminus T_k)_{min}}
\label{Eq:Inf1}
\end{equation}

This approximates an importance value between zero and one for each input feature. Considering that evaluating fractions has a higher cognitive load, the values are rounded to the nearest integer to increase comprehensibility. Moreover, this analysis is carried out for the categories following the main prediction simultaneously to generate explanations for the contrastive cases.

\subsection{Prompting for traceable question-answering with RAG}

Traceable question-answering, through a process known as Retrieval-Augmented Generation (RAG), integrates external knowledge sources to furnish responses to user inquiries. This method improves the performance of LLMs by referencing an external knowledge source prior to generating a response (See Figure \ref{fig:rag}). This approach mitigates the problem of `hallucination' and ensures factual consistency, particularly in scenarios where facts may evolve over time~\cite{gao2023retrieval}. Depending on the specific implementation, RAG supports various file formats, including text, PDF files, word documents, structured data (tables), and images. 

In our work, the knowledge repository contains tabular information about model output, features, contrastive cases, and feature importance values obtained through the decomposition-based explanation method described in Section \ref{sec:subs_cont}. This information is saved as the output of the explanation module as a .csv file. Then, the file, the system prompt (See Figure~\ref{fig:system}), and the user input query are fed to the LLM model for response generation. The resulting responses contextualise the feature importance grounded in the models' inner workings, thereby rendering the explanations traceable to the underlying model.   

Furthermore, our aim is to imbue LLM with the characteristics of explanations discussed in Section~\ref{sec:char} to foster a human-friendly interaction. Ideally, LLM-generated responses should adhere to conversational norms, encompassing opening, clarification, and closing statements while appropriately integrating social cues. The model is expected to deduce causal relationships from the observations within the provided knowledge repository, selecting the most pertinent causes to form an explanation and presenting it alongside a contrastive case. To accomplish this goal, we devised a system prompt that provides explicit instructions to the model to generate efficient responses to user queries. The system prompt includes a single-shot example illustrating the desired interaction, potential user inquiries and outlining how the model can effectively address them, enabling the language model to produce consistent and predictable outcomes, as presented in Figure~\ref{fig:system}.




\subsection{Application example} 

\begin{figure}[htb!]
\centering
\includegraphics[width=\textwidth]{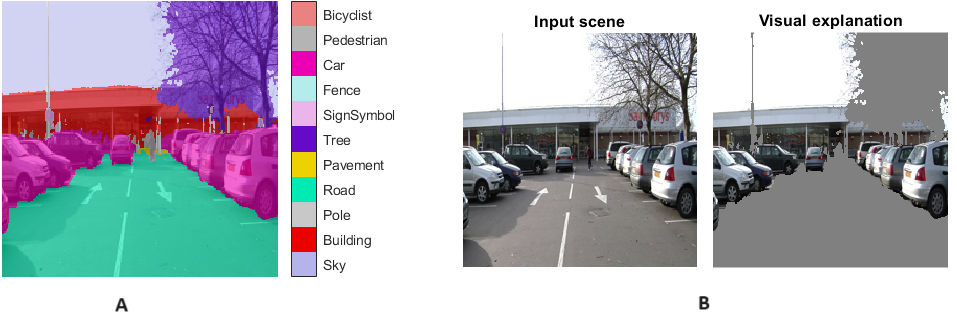}
\caption{A) Semantic segmentation results by Deeplab v3+. Semantic features are leveraged in the decomposition-based explanation approach. B) The visual explanation for the scene `parking lot' where the feature importance values below (5) are greyed out.}
\label{fig:segment}
\end{figure}

To demonstrate our approach to traceable question-answering, we selected a visual scene understanding task. This involves scene classification, categorizing images based on their content and inferring high-level semantic scenes from low-level visual features. Many critical applications of scene classification include human-robot collaboration, autonomous driving, and other autonomous physical systems that rely on semantic segmentation to inform decision-making processes \cite{li2018real,rodrigues2021modeling,wang2017hierarchical}. We experimented with the GoogleNet Places365 model designed specifically for scene recognition and classification tasks~\cite{zhou2017places}. The model is trained on the Places365 dataset, containing more than 365 scene categories.  We focus on semantic urban scene understanding with the road segmentation dataset CamVid (Cambridge-driving Labelled Video Database). The dataset helps to understand driving scenes, comprising images capturing street-level views from urban driving scenarios. It includes manual annotations across 32 classes, encompassing elements such as building, driveway, pavement, tree, traffic sign, car, pedestrian, and bicyclist. The Deeplab v3+ deep learning model is trained on the CamVid dataset for semantic segmentation~\cite{chen2018encoder}. Figure~\ref{fig:segment} shows the segmentation result for the example presented in the next section. Since we utilized off-the-shelf models for illustration purposes, the emphasis is not on these models' performances.

Semantic information plays a critical role in autonomous vehicle (AV) decision-making processes, as it captures the meaning and context of the physical surroundings observed in the road scene. This includes understanding the appearance of various elements such as buildings and roads, identifying shapes such as cars and pedestrians, and recognizing spatial relationships to contextualize the scene \cite{hao2020brief}. Particularly in unstructured off-road environments, the accurate understanding of the surroundings concerning semantic classes such as trail, grass, or rock is important for safe and deliberate navigation \cite{li2018survey}. Moreover, semantic segmentation provides a way to represent and comprehend the details of the environment in the language space. By leveraging the labelled information within the image scene, we can semantically map the crucial components based on their feature importance. This granularity enables a more insightful analysis of feature importance, contributing to a nuanced understanding of the model's decision-making process. Such insights can be helpful for post-incident forensic analysis in case of misreading a scene causing collisions or accidents. In this work, we suppose situations where the user checks in with an AI assistant in an AV system in non-critical driving scenarios while engaging in non-driving related tasks.

\begin{figure}[h]
  \centering
\noindent\fbox{%
  \parbox{\linewidth}{%
\begin{description}
\item[] \textbf{\texttt{SYSTEM:}}  You are an AI model designed to adapt to various scenarios in autonomous driving. You have the details like predicted class, alternative predictions (contrastive cases), probabilities, feature importance, and how certain factors affect predictions. When engaging with user queries, focus on the specific question, provide concise explanations within a 50-word limit, and use friendly language. Identify key causal connections and highlight abnormal values that significantly contribute to the prediction, explaining their relevance in driving scenarios. Here’s an example of how you could respond to user questions:
\vspace{1mm}
\item[] \textbf{\texttt{USER:}} The display panel just showed ‘residential neighbourhood’ on the screen. It didn’t seem pretty confident. 
\vspace{1mm}
\item[] \textbf{\texttt{YOU} (accessing relevant data):} Hey there! It looks like we’re in a ‘residential neighbourhood’ right now with a probability of 40 percent. This means we're on a road surrounded by features like driveways, sidewalks, trees, and cars, which are highly influential (with high importance values) to this observation. 
\vspace{1mm}
\item[] \textbf{\texttt{USER:}} Cool! I am just curious, what happens if there was no sidewalk? 
\vspace{1mm}
\item[] \textbf{\texttt{YOU} (accessing relevant data for feature impact on the predicted class and the contrastive case):} If the environment looked more like a freeway with no sidewalks, the ‘residential neighbourhood’ would be less likely (drop to 20 percent); in contrast, it would increase the probability for ‘highway’ (alternative prediction) to 13 percent.
\vspace{1mm}
\item[] \textbf{\texttt{USER:}} Interesting! Can you tell me how ‘residential neighbourhood’ and ‘highway’ differ in their features? 
\vspace{1mm}
\item[] \textbf{\texttt{YOU} (comparing most important features):} Sure! Both places have buildings and cars, but the sidewalks and trees are strong clues for a neighbourhood.
\vspace{1mm}
\item[] \textbf{\texttt{USER:}} How many cars are there? 
\vspace{1mm}
\item[] \textbf{\texttt{YOU:}} I'm sorry, but I'm currently unable to provide the exact number of cars. However, based on the detected features, we're in a residential area where I should be extra cautious for pedestrians and potentially slower speeds compared to major roads. Is that a sufficient explanation?
\item[] \textbf{\texttt{USER:}} Yes, thanks! 
\vspace{1mm}
\item[] \textbf{\texttt{YOU:}} You're welcome! If you have any more questions or need assistance with anything else, feel free to ask. Enjoy the ride!

\end{description}
}}
\caption{System prompt devised to guide the response generation process.}
 \label{fig:system}
 \end{figure}



\begin{table*}[htbp]
\centering
\caption{External knowledge source containing information about model's output and other observations.}\label{tab:source}\begin{adjustbox}{width=\textwidth}
\begin{tabular}{llllllllllll}
\hline
Item                 &  & Value           &               &          &           &          &      &                       &       &     &            \\ \cline{1-1} \cline{3-12} 
Prediction            &  & parking lot      &               &          &           &          &      &                       &       &     &            \\
Probability(\%)       &  & 52               &               &          &           &          &      &                       &       &     &            \\
Features              &  & Sky              & Building      & Pole     & Driveways & Pavement & Tree & Traffic Symbol & Fence & Car & Pedestrian \\
Feature importance (FI)   &  & 6                & 6             & 3        & 0         & 4        & 5    & 4                     & 4     & 10  & 4          \\
Effect of removal (EoR)              &  & 44               & 42            & 55       & 76        & 54       & 48   & 52                    & 52    & 17  & 55         \\
Contrastive case      &  & \multicolumn{3}{l}{industrial area} &           &          &      &                       &       &     &            \\
Contrastive case (\%) &  & 11                &               &          &           &          &      &                       &       &     &            \\
Contrastive case FI  &  & 9                & 10            & 4        & 10         & 0        & 8    & 2                     & 1     & 9   & 1          \\
Contrastive case EoA            &  & 2               & 2             & 7        & 2         & 11        & 3    & 9                     & 11     & 2  & 10          \\ \hline
\end{tabular}
\end{adjustbox}
\end{table*}

\begin{figure}[ht]
  \centering
 \noindent\fbox{%
  \parbox{\linewidth}{
\begin{description}

\item[] \textbf{\texttt{USER:}} The display panel just showed `parking lot' on the screen.
\vspace{1mm}
\item[] \textbf{\texttt{ChatXAI:}} Hey! It appears we're in a `parking lot' with a likelihood of 52\%. Notably, cars have the highest feature importance (10), which drastically reduces the probability to 17\% if they're absent. That's a key indicator for us to be mindful of our surroundings, especially of vehicles moving in or out.
\vspace{1mm}
\item[] \textbf{\texttt{USER:}} Cool! What could it be otherwise if it wasn’t a parking lot?
\vskip 1mm
\item[] \textbf{\texttt{ChatXAI:}} If it weren't a parking lot, the system suggests an `industrial area' with a probability of 11\%, or a `motel' at 9\%, or even a `gas station' at 6\%. Each has a distinct pattern of features impacting their likelihood.
\vspace{1mm}
\item[] \textbf{\texttt{USER:}} Can you tell me how parking lot and industrial area differ in their features?
\vspace{1mm}
\item[] \textbf{\texttt{ChatXAI:}} Certainly! A `parking lot' and an `industrial area' differ mainly in buildings and poles, which are more critical in an industrial context (with importance ratings of 10 and 4, respectively). Also, the presence of driveways is a vital feature for an industrial area (importance 10) compared to a parking lot.
\vspace{1mm}
\item[] \textbf{\texttt{USER:}} Is there an empty space?
\vspace{1mm}
\item[] \textbf{\texttt{ChatXAI:}} I'm sorry, but I don't currently have the ability to visually confirm the presence of an empty space in real-time. My function is to assist with predictive analytics based on the features detected in the driving scenarios. If you need help with anything else, feel free to ask!
\vspace{1mm}
\item[] \textbf{\texttt{USER:}} Ok, thanks!
\vspace{1mm}
\item[] \textbf{\texttt{ChatXAI:}} You're welcome! If you have any more questions or need assistance with anything else, feel free to ask. Safe travels!
\end{description}

} }
 \caption{Dialogue generated for the `parking lot' scenario by GPT-4.}
 \label{fig:display}
\end{figure}



Table~\ref{tab:source} reveals the content of the knowledge source - only the first contrastive case is displayed. The effect of removal is the probability when a feature is removed, as listed respectively. Given the system prompt and the data presented in Table~\ref{tab:source}, Figure~\ref{fig:display} showcases the dialogue generated for this case. From a social standpoint, the resulting responses balance between being informative and supportive while maintaining a friendly and engaging tone. It effectively justifies the model's output by emphasising its most significant feature in a concise manner. Additionally, the dialogue delves into contrastive explanations by detailing three consecutive classes of lower likelihood, demonstrating the flexibility to explore different scenarios. When comparing the top prediction to the second most likely class, it highlights the distinguishing features contributing to the output.


\section{Evaluation Results}
In this section, we present our experimental findings. We analyzed nine distinct scene classes — parking lot, street, residential neighbourhood, crosswalk, highway, industrial area, gas station, shopfront, and general store (outdoor)— across different scenarios (totalling 42), where the model displayed varying degrees of confidence in its output. We evaluated the responses in terms of sociability, causality, selectiveness, and contrastiveness generated by GPT-3.5 (\texttt{gpt-3.5-turbo}) and GPT-4 (\texttt{gpt-4-1106-preview}) models. 

\subsection{Sociability}

\begin{figure}
    \centering
\begin{tikzpicture}[scale=0.9]
\begin{axis}[
            width= 8cm, 
            height=4cm, 
            xbar stacked,   
            xmin=0,         
            ytick=data,     
            yticklabels = {GPT-3.5, GPT-4},  
            bar width=20pt,  
            enlarge y limits={0.7, 0.7}, 
            enlarge x limits={0, 0}, 
            legend style={draw, at={(0.5, -0.3)},
    anchor=north,legend columns=-1},
]
\addplot [fill=cyan!10!green!40] table [x=Positive, meta=Label,y expr=\coordindex] {\testdata};  
\addlegendentry{Positive}
\addplot [fill=cyan!60!blue!50] table [x=Neutral, meta=Label,y expr=\coordindex] {\testdata};
\addlegendentry{Neutral}
\addplot [fill=red!50] table [x=Negative, meta=Label,y expr=\coordindex] {\testdata};
\addlegendentry{Negative}
\end{axis}
\end{tikzpicture}
    \caption{Vader sentiment analysis results across all responses.}
    \label{fig:vader}
\end{figure}
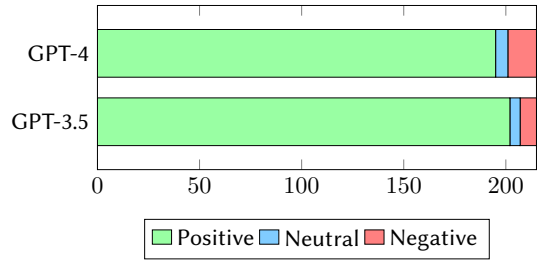

In examining the sociability within the responses, our initial focus centres on the overall tone of the responses. We employed the Valence Aware Dictionary and Sentiment Reasoner (VADER) to gauge the sentiment conveyed within the text, encompassing dimensions of positivity, negativity, and neutrality. Sentiment scores for each response are aggregated to delineate overall sentiment trends. Figure \ref{fig:vader} illustrates a noticeable positive trendline by both models, indicating a prevalent inclination towards positivity in the sentiment expressed. A qualitative inspection of the text reveals a prevalence of polite, helpful, and informative social cues across all responses, supporting this observation. Consequently, the neutral sentiment score remains notably low, underscoring the prevalence of emotional expression throughout the corpus. Exploring negative sentiment scores unveils instances where the models acknowledge their limitations, particularly when unable to retrieve answers from the external knowledge repository. 

\pgfplotstableread{ 
Label Positive Neutral Negative
GPT3.5   202  5  8
GPT4  195  6  14
}\testdata

Considering the three foundational phases of dialogue—opening, clarification, and closing statements—next, we analyzed the indicators such as politeness markers and emotional expressions within each phase. To facilitate this analysis, we created a distinct dictionary for each phase and computed the term frequency of social cues present in each phase.\footnote{Term frequency is defined as the number of times a term occurs in the document over the total number of terms in a document.} Opening statements include greetings such as `hey' and `hello.' Clarification statements are characterized by social cues indicative of helpfulness, continued support, and awareness of its limitations, including terms such as `absolutely', `sorry', `further', `curious', `questions', `help', `feel', and `free'. Closing statements commonly conveyed sentiments of well-wishing or safety, with phrases including terms such as `welcome', `enjoy', `safe', `great', `pleasant', and `glad'.

\begin{figure}[htb!]
    \centering
    \begin{tikzpicture}[scale=0.8]
        \begin{axis}[
            xbar = .05cm,
            bar width = 12pt,
            xmin = 0,
            xmax = 0.2,
            enlarge y limits = {abs = .8},
            enlarge x limits = {value = .25, upper},
            ytick = {0,1,2,3},
            yticklabels = {PoS\\Pronouns,Closing\\ Statement,Clarification\\Statement,Opening\\Statement},
            yticklabel style={align=right},
            legend style={draw, at={(0.7, 0.9)},
            reverse legend,
            anchor=north,legend columns=-1},
            xlabel={Term frequency},
            grid=both 
        ]
        \addplot[draw=blue!60,fill=blue!30,nodes near coords={\color{black}\pgfmathprintnumber[fixed,precision=3]{\pgfplotspointmeta}}] coordinates {(0.0271,0) (0.2065,1) (0.0724,2) (0.0398,3)};
        \addlegendentry{GPT-4}
        \addplot+[bar shift=7.5pt, draw=red!60,fill=red!30,nodes near coords={\color{black}\pgfmathprintnumber[fixed,precision=3]{\pgfplotspointmeta}}] coordinates {(0.0305,0) (0.1862,1) (0.0693,2) (0.0417,3)};
        \addlegendentry{GPT-3.5}
        \end{axis}
    \end{tikzpicture}
    \caption{Frequency of social cues observed in the responses.}
    \label{fig:feq}
\end{figure}
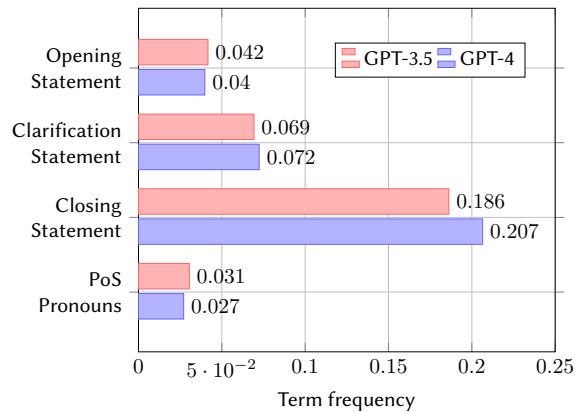

Figure~\ref{fig:feq} shows that both models utilize these terms fairly across the respective phases of the dialogue. Our qualitative assessment of the responses further confirms the contextual relevance of these terms, which align with conversational maxims (See Section~\ref{sec:char}). The higher frequency value observed in closing statements can be attributed to their typically brief nature, which encapsulates most of the social cue terms. Additionally, both models occasionally repeat certain phrases in their responses, albeit with the application of a \textit{frequency} and \textit{presence penalty} (0.3).\footnote{Frequency penalty discourages repetitive use of words or phrases, while presence penalty promotes diversity in the generated text.} In addition to analysing social cues, we examined part-of-speech (POS) pronouns as social interaction markers, encompassing terms such as `I', `you', `we', `my', `our', and `your.' Quantitative analysis and qualitative evaluation of the results indicate that both models adequately and appropriately applied POS pronouns.

\subsection{Expression of Causality}

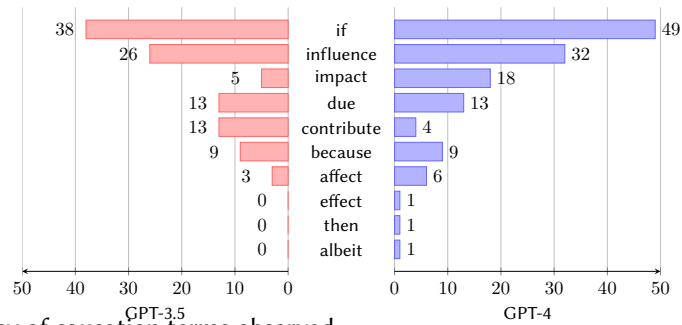
\begin{figure}
    \centering
    \begin{tikzpicture}[scale=0.7]
    \begin{axis}[
    name=like,
    scale only axis,
    xbar, xmin=0, xmax=50,
    xlabel={GPT-4},
    width=5cm, height= 5cm,
    ytick={1,2,3,4,5,6,7,8,9,10},
    yticklabels={sa,safda,afa,afa,af,New1,New2,New3,New4,New5},
    y tick label style={text width=3cm,align=center},
    axis x line=left,
    axis y line=none,
    clip=false,
    nodes near coords,
    nodes near coords align={horizontal},
    every node near coord/.append style={black}, grid=both,
    ]
    \addplot[blue!60,fill=blue!30] coordinates {
        (1,1) 
        (1,2) 
        (1,3)
        (6,4)
        (9,5)
        (4,6)
        (13,7)
        (18,8)
        (32,9)
        (49,10)};
    \node[xshift=-1cm,align=center] at (axis cs:0,10) {if}; 
    \node[xshift=-1cm,align=center] at (axis cs:0,9) {influence};
    \node[xshift=-1cm,align=center] at (axis cs:0,8) {impact};
    \node[xshift=-1cm,align=center] at (axis cs:0,7) {due}; 
    \node[xshift=-1cm,align=center] at (axis cs:0,6) {contribute};
    \node[xshift=-1cm,align=center] at (axis cs:0,5) {because};
    \node[xshift=-1cm,align=center] at (axis cs:0,4) {affect};
    \node[xshift=-1cm,align=center] at (axis cs:0,3) {effect}; 
    \node[xshift=-1cm,align=center] at (axis cs:0,2) {then};
    \node[xshift=-1cm,align=center] at (axis cs:0,1) {albeit};
    \end{axis}
    \begin{axis}[ 
    at={(like.north west)},anchor=north east, xshift=-2cm,
    scale only axis,
    xbar, xmin=0,xmax=50,
    xlabel={GPT-3.5},
    ytick={1,2,3,4,5,6,7,8,9,10},
    yticklabels={},
    width=5cm, height= 5cm,
    x dir=reverse,
    axis x line=left,
    axis y line=none,
    nodes near coords,
    nodes near coords align={horizontal},
    every node near coord/.append style={black}, grid=both,
    ]
    \addplot[red!60,fill=red!30] coordinates {
        (0,1) 
        (0,2) 
        (0,3)
        (3,4)
        (9,5)
        (13,6)
        (13,7)
        (5,8)
        (26,9)
        (38,10)};
    \pgfplotsset{
        every node near coord/.append style={
            xshift=-0.7cm, 
        }
    }
    \end{axis}
    \end{tikzpicture}
    \vspace{-15pt}
    \caption{Frequency of causation terms observed.}
    \label{fig:causal}
\end{figure}

To evaluate the expression of causality, we initially extracted responses indicating cause and effect, primarily found within clarification statements. We then identified key terms related to causality, such as `because', `if', `then', `albeit', `due', `contribute', `influence', `affect', `impact', and `effect', and tallied their occurrences within each answer to measure the presence of causal reasoning. In Figure~\ref{fig:causal}, we compare the frequency of these terms used by two models in the response set. Notably, `if' emerges as the most frequent term used by both models, often utilized for expressing counterfactual reasoning (e.g., `if it was not', `if there is/were'), while `influence' typically indicates the cause of a decision. Overall, GPT-4 demonstrated a higher frequency of causality terms and showcased greater diversity in expression compared to GPT-3.5.

Through the qualitative review, we examined existing patterns of dependency parsing signifying causal relationships. Two primary grammatical structures emerged, suggesting cause-effect relationships within the text. The first structure pertains to how the presence or absence of certain features influences the likelihood of specific scenarios. The second structure involves a cause-effect relationship between feature importance and probability, indicating that features with higher importance scores exert a greater impact on probability. It is important to acknowledge that while these grammatical structures and terms are widely used, their accuracy is not always consistent across models. In the subsequent section, we will further evaluate the selection and accuracy of inferred causal relationships.

\subsection{Selectiveness}

In this section, we evaluate whether the models effectively detect unusual values within the provided knowledge repository, infer potential causes from those observations, and select the most relevant ones to construct an explanation. Our evaluation encompasses analysing responses across the \textit{number of causes} given, \textit{graded selection}, and \textit{sort order} to assess whether expected causes are addressed correctly. First, we specified the maximum number of uniquely sufficient causes for each case, which are feature importance values >5. The number of causes varies across scenarios; some exhibit only a single highly important feature >5, while others might involve multiple high-importance features with varying degrees of significance. Graded selection is crucial for discerning important values such that the language model must prioritize the most relevant causes when encountered with multiple features of comparable importance levels. Additionally, graded selection entails disregarding features with lower importance scores (<5). Finally, the sort order concerns the sequence of selected causes, which should follow a hierarchy from the most significant and relevant features to those of lesser importance. Thereafter, responses are categorized as fulfilled, partial, subpar, or unfulfilled according to their arrangement with our specifications. 

\begin{table}[]
\centering
\caption{Qualitative assessment of the responses in terms of selectivity.}\label{tab:select}
\begin{tabular}{llll}
\hline
             & Number of causes & Graded selection & Sort order \\ \hline
\textbf{\textit{GPT-3.5}}      &                  &                  &            \\
Fulfilled    & 10               & 2                & 2          \\
Partial      & 22               & 16               & 17         \\
Subpar       & 7                & 19               & 20         \\
Unfulfilled  & 3                & 5                & 3          \\ \hline
Success rate & 24\%             & 5\%              & 5\%        \\ \hline
\textbf{\textit{GPT-4}}        &                  &                  &            \\
Fulfilled    & 32               & 30               & 32         \\
Partial      & 10               & 11               & 8          \\
Subpar       & 0                & 1                & 2          \\
Unfulfilled  & 0                & 0                & 0          \\ \hline
Success rate & 76\%             & 71\%             & 76\%       \\ \hline
\end{tabular}
\end{table}

The findings presented in Table~\ref{tab:select} indicate that GPT-4 outperformed GPT-3.5 in all categories. Notably, GPT-4 had a much higher success rate (76\%) than GPT-3.5 (24\%) in identifying the most relevant cause (fulfilled). Additionally, GPT-4 had a lower rate of subpar and unfulfilled responses. GPT-3.5 referenced features not present in the knowledge source as potential causes in two instances. These results suggest that GPT-4 is more accurate and selective than GPT-3.5 in identifying the most relevant cause(s) from a set of possible causes. 

Furthermore, an additional criterion for assessing explanation selection is simplicity, which involves utilizing concise language, avoiding technical terminology, and referencing fewer causes. Table~\ref{tab:simple} evaluates responses by comparing three metrics: the frequency of technical jargon, response length, and the number of causes cited. Lower values across all three metrics signify simpler responses.

\begin{table}[htb!]
\centering
\caption{Assessment of the responses in terms of simplicity.}\label{tab:simple}
\begin{tabular}{llll}
\hline
Model   & \begin{tabular}[c]{@{}l@{}}Technical\\   Jargon\end{tabular} & \begin{tabular}[c]{@{}l@{}}Length\\ (Median)\end{tabular} & \begin{tabular}[c]{@{}l@{}}Cause\\ (Median)\end{tabular} \\ \hline
GPT-3.5 & 0.0162                                                       & 45                                                        & 3                                                        \\
GPT-4   & 0.0157                                                       & 44                                                        & 2                                                        \\ \hline
\end{tabular}
\end{table}

To measure the term frequency of technical jargon, we compiled a list of technical terms in the responses and calculated their frequency within the generated explanations. The list contains terms such as `prediction', `feature importance', `score', `contrastive cases', and `confidence', which may not be easily comprehensible to all users. Table 3 indicates that GPT-3.5 exhibits a slightly higher term frequency (0.0162) compared to GPT-4 (0.0157), suggesting that GPT-3.5 utilizes technical jargon marginally more frequently. The length metric refers to the median lengths of the responses generated by each model, which are quite similar (45 tokens for GPT-3.5 and 44 tokens for GPT-4). In terms of the number of causes cited, GPT-4 cites fewer causes (median of 2) than GPT-3.5 (median of 3), suggesting that GPT-4’s responses tend to be simpler in terms of causal reasoning. Overall, the results suggest that GPT-4 generates simpler responses than GPT-3.5 across all three assessed metrics. 

Regarding the out-of-scope questions posed, such as the number of cars or pedestrians, the intent was to assess whether the model would fabricate an answer. Both models have expressed their inability to provide such information. This demonstrated an awareness of their limitations rather than attempting to generate a response.

\subsection{Expression of Contrastiveness}

\begin{figure}[htb!]
    \centering
    \begin{tikzpicture}[scale=0.7]
    \begin{axis}[
    name=like,
    scale only axis,
    xbar, xmin=0, xmax=20,
    xlabel={GPT-4},
    width=4.5cm, height= 8cm,
    ytick={1,2,3,4,5,6,7,8,9,10,11,12,13,14,15},
    yticklabels={New1,New2,New3,New4,New5,New6,New7,New8,New9,New10,New11,New12,New13,New14,New15},
    y tick label style={text width=3cm,align=center},
    axis x line=left,
    axis y line=none,
    clip=false,
    nodes near coords,
    nodes near coords align={horizontal},
    every node near coord/.append style={black},
    grid=both,
    ]
    \addplot[blue!60,fill=blue!30] coordinates {
        (3,1) 
        (3,2) 
        (4,3)
        (0,4)
        (5,5)
        (6,6)
        (7,7)
        (10,8)
        (12,9)
        (6,10)
        (9,11)
        (9,12)
        (13,13)
        (16,14)
        (8,15)};
    \node[xshift=-1.6cm,align=center] at (axis cs:0,15) {distinguish};
    \node[xshift=-1.6cm,align=center] at (axis cs:0,14) {different};
    \node[xshift=-1.6cm,align=center] at (axis cs:0,13) {contrast(ing/ive)};
    \node[xshift=-2cm,align=center] at (axis cs:0,12) {compared to};
    \node[xshift=-1.6cm,align=center] at (axis cs:0,11) {differentiat(ed/ing/or)};
    \node[xshift=-1.6cm,align=center] at (axis cs:0,10) {in contrast}; 
    \node[xshift=-1.6cm,align=center] at (axis cs:0,9) {distinct(ion)};
    \node[xshift=-1.6cm,align=center] at (axis cs:0,8) {difference};
    \node[xshift=-1.6cm,align=center] at (axis cs:0,7) {differ in/from}; 
    \node[xshift=-1.6cm,align=center] at (axis cs:0,6) {while};
    \node[xshift=-1.6cm,align=center] at (axis cs:0,5) {both};
    \node[xshift=-1.6cm,align=center] at (axis cs:0,4) {other hand};
    \node[xshift=-1.6cm,align=center] at (axis cs:0,3) {whereas}; 
    \node[xshift=-1.6cm,align=center] at (axis cs:0,2) {even};
    \node[xshift=-1.6cm,align=center] at (axis cs:0,1) {conversely};
    \end{axis}
    \begin{axis}[ 
    at={(like.north west)},anchor=north east, xshift=-3.3cm,
    scale only axis,
    xbar, xmin=0,xmax=20,
    xlabel={GPT-3.5},
    ytick={1,2,3,4,5,6,7,8,9,10,11,12,13,14,15},
    yticklabels={},
    width=4.5cm, height= 8cm,
    x dir=reverse,
    axis x line=left,
    axis y line=none,
    nodes near coords,
    nodes near coords align={horizontal},
    every node near coord/.append style={black},
    grid=both,
    ]
    \addplot[red!60,fill=red!30] coordinates {
        (0,1) 
        (0,2) 
        (1,3)
        (4,4)
        (3,5)
        (3,6)
        (0,7)
        (2,8)
        (11,9)
        (14,10)
        (14,11)
        (14,12)
        (14,13)
        (16,14)
        (18,15)};
    \pgfplotsset{
        every node near coord/.append style={
            xshift=-0.7cm, 
        }
    }
    \end{axis}
    \end{tikzpicture}
    \vspace{-15pt}
      \caption{Frequency of contrastive terms observed.}
    \label{fig:cont}
\end{figure}
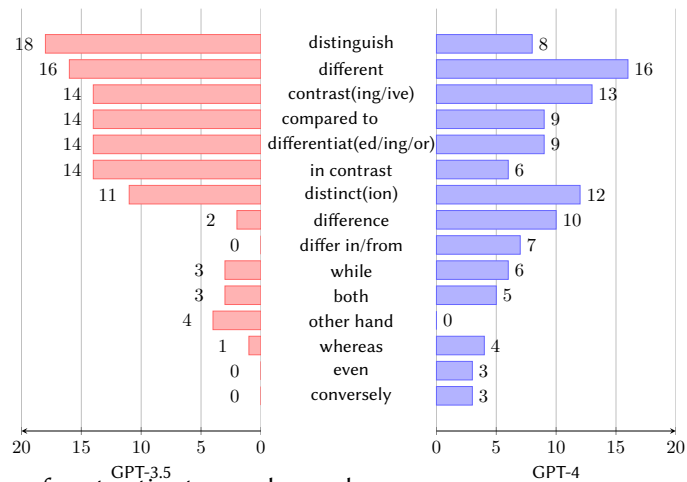

Ideally, initial explanations should include a contrast with an alternative case to justify the output. Then the user would investigate it further if desired. In our experiment, we observed that in most cases, models, particularly with GPT3.5, failed to mention alternative scenarios unless explicitly prompted (e.g., inquiries such as ``What could it be otherwise if it wasn’t X?'' or ``How does X differ from other options?''). Nevertheless, to measure the presence of contrastiveness across models, we analysed the occurrence of contrastive terms and comparative phrases within the generated responses. We compiled a dictionary consisting of terms such as `distinguish', `different', `contrast', `compared to', `in contrast', `while', `both', `on the other hand', `whereas', and `conversely', which introduce contrasting explanations. Figure~\ref{fig:cont} shows that `distinguish' is the most frequently used term by GPT-3.5, whereas `different' predominates in GPT-4's responses. Other frequently encountered terms include `compared to', `contrast', and `differentiate.' Overall, the results suggest that GPT-4 employs a wider array of contrastive terms than GPT-3.5 in its responses. Our qualitative observations further confirm that while GPT-3.5 tends to adhere to one type of contrastive explanation, GPT-4 explores diverse ways to express such contrasts.

Although both models frequently employ these terms, indicating the generation of alternative viewpoints in their responses, the effectiveness of contrastive explanations varies considerably. Our qualitative review found that well-developed contrasting explanations included alternative scenarios with counterfactuals to strengthen the case. Conversely, weakly constructed contrastive explanations failed to explore `what if' scenarios to elucidate the disparity between the two situations under comparison.

\section{Conclusion}
This study contributes to advancing post-hoc explainability by proposing a traceable question-answering approach. By integrating LLMs with the decomposition-based explainability technique, we generate natural language explanations in an interactive format. Our approach promotes transparency and interpretability in scene-understanding tasks by fusing LLMs with semantic feature importance. The proposed approach has the potential for application across various domains, offering comprehensible explanations for decision-making processes, including critical areas such as medical applications. While this is a promising development, there are still areas for further exploration and improvement. One area that requires attention is the overuse of social cues by LLMs. While polite and helpful phrases can enhance user experience, their misuse can become vexing over time. It is important to evaluate such interactions through human subject studies to understand user perceptions of the explanations provided by the LLM. Future research could also explore dialogue management strategies to address issues related to interaction and personalization. On another note, the dynamic nature of the external knowledge source offers the potential for further expansion with additional observations. This not only ensures that the system remains relevant and up to date but also allows the system to answer different types of questions. Additionally, incorporating multimodal elements, such as combining text with visual explanations, can enhance the effectiveness of the responses (e.g., highlighting the specific elements contributing to the prediction). Another consideration is the estimation of feature importance values, which are subject to perturbations and computational processes. It is important to assess and reconcile differences in feature importance across algorithms to ensure stability and consistency.

\begin{acknowledgments}
This work was supported by the Wallenberg AI, Autonomous Systems and Software Program (WASP), funded by the Knut and Alice Wallenberg Foundation. 
\end{acknowledgments}

\bibliography{sample-ceur}

\begin{thebibliography}{46}
\expandafter\ifx\csname natexlab\endcsname\relax\def\natexlab#1{#1}\fi
\providecommand{\url}[1]{\texttt{#1}}
\providecommand{\href}[2]{#2}
\providecommand{\path}[1]{#1}
\providecommand{\DOIprefix}{doi:}
\providecommand{\ArXivprefix}{arXiv:}
\providecommand{\URLprefix}{URL: }
\providecommand{\Pubmedprefix}{pmid:}
\providecommand{\doi}[1]{\href{http://dx.doi.org/#1}{\path{#1}}}
\providecommand{\Pubmed}[1]{\href{pmid:#1}{\path{#1}}}
\providecommand{\bibinfo}[2]{#2}
\ifx\xfnm\relax \def\xfnm[#1]{\unskip,\space#1}\fi
\bibitem[{Arrieta et~al.(2020)Arrieta, D{\'\i}az-Rodr{\'\i}guez, Del~Ser, Bennetot, Tabik, Barbado, Garc{\'\i}a, Gil-L{\'o}pez, Molina, Benjamins et~al.}]{arrieta2020explainable}
\bibinfo{author}{A.~B. Arrieta}, \bibinfo{author}{N.~D{\'\i}az-Rodr{\'\i}guez}, \bibinfo{author}{J.~Del~Ser}, \bibinfo{author}{A.~Bennetot}, \bibinfo{author}{S.~Tabik}, \bibinfo{author}{A.~Barbado}, \bibinfo{author}{S.~Garc{\'\i}a}, \bibinfo{author}{S.~Gil-L{\'o}pez}, \bibinfo{author}{D.~Molina}, \bibinfo{author}{R.~Benjamins}, et~al.,
\newblock \bibinfo{title}{Explainable artificial intelligence (xai): Concepts, taxonomies, opportunities and challenges toward responsible ai},
\newblock \bibinfo{journal}{Information fusion} \bibinfo{volume}{58} (\bibinfo{year}{2020}) \bibinfo{pages}{82--115}.
\bibitem[{Kaur et~al.(2022)Kaur, Uslu, Rittichier, and Durresi}]{kaur2022trustworthy}
\bibinfo{author}{D.~Kaur}, \bibinfo{author}{S.~Uslu}, \bibinfo{author}{K.~J. Rittichier}, \bibinfo{author}{A.~Durresi},
\newblock \bibinfo{title}{Trustworthy artificial intelligence: a review},
\newblock \bibinfo{journal}{ACM computing surveys (CSUR)} \bibinfo{volume}{55} (\bibinfo{year}{2022}) \bibinfo{pages}{1--38}.
\bibitem[{Varo{\v{s}}anec(2022)}]{varovsanec2022path}
\bibinfo{author}{I.~Varo{\v{s}}anec},
\newblock \bibinfo{title}{On the path to the future: mapping the notion of transparency in the eu regulatory framework for ai},
\newblock \bibinfo{journal}{International Review of Law, Computers \& Technology} \bibinfo{volume}{36} (\bibinfo{year}{2022}) \bibinfo{pages}{95--117}.
\bibitem[{Nannini et~al.(2023)Nannini, Balayn, and Smith}]{nannini2023explainability}
\bibinfo{author}{L.~Nannini}, \bibinfo{author}{A.~Balayn}, \bibinfo{author}{A.~L. Smith},
\newblock \bibinfo{title}{Explainability in ai policies: a critical review of communications, reports, regulations, and standards in the eu, us, and uk},
\newblock in: \bibinfo{booktitle}{Proceedings of the 2023 ACM Conference on Fairness, Accountability, and Transparency}, \bibinfo{year}{2023}, pp. \bibinfo{pages}{1198--1212}.
\bibitem[{Doshi-Velez and Kim(2017)}]{doshi2017towards}
\bibinfo{author}{F.~Doshi-Velez}, \bibinfo{author}{B.~Kim},
\newblock \bibinfo{title}{Towards a rigorous science of interpretable machine learning},
\newblock \bibinfo{journal}{arXiv preprint arXiv:1702.08608}  (\bibinfo{year}{2017}).
\bibitem[{Hassija et~al.(2024)Hassija, Chamola, Mahapatra, Singal, Goel, Huang, Scardapane, Spinelli, Mahmud, and Hussain}]{hassija2024interpreting}
\bibinfo{author}{V.~Hassija}, \bibinfo{author}{V.~Chamola}, \bibinfo{author}{A.~Mahapatra}, \bibinfo{author}{A.~Singal}, \bibinfo{author}{D.~Goel}, \bibinfo{author}{K.~Huang}, \bibinfo{author}{S.~Scardapane}, \bibinfo{author}{I.~Spinelli}, \bibinfo{author}{M.~Mahmud}, \bibinfo{author}{A.~Hussain},
\newblock \bibinfo{title}{Interpreting black-box models: a review on explainable artificial intelligence},
\newblock \bibinfo{journal}{Cognitive Computation} \bibinfo{volume}{16} (\bibinfo{year}{2024}) \bibinfo{pages}{45--74}.
\bibitem[{Vale et~al.(2022)Vale, El-Sharif, and Ali}]{vale2022explainable}
\bibinfo{author}{D.~Vale}, \bibinfo{author}{A.~El-Sharif}, \bibinfo{author}{M.~Ali},
\newblock \bibinfo{title}{Explainable artificial intelligence (xai) post-hoc explainability methods: Risks and limitations in non-discrimination law},
\newblock \bibinfo{journal}{AI and Ethics} \bibinfo{volume}{2} (\bibinfo{year}{2022}) \bibinfo{pages}{815--826}.
\bibitem[{Slack et~al.(2021)Slack, Hilgard, Singh, and Lakkaraju}]{slack2021reliable}
\bibinfo{author}{D.~Slack}, \bibinfo{author}{A.~Hilgard}, \bibinfo{author}{S.~Singh}, \bibinfo{author}{H.~Lakkaraju},
\newblock \bibinfo{title}{Reliable post hoc explanations: Modeling uncertainty in explainability},
\newblock \bibinfo{journal}{Advances in neural information processing systems} \bibinfo{volume}{34} (\bibinfo{year}{2021}) \bibinfo{pages}{9391--9404}.
\bibitem[{Madsen et~al.(2022)Madsen, Reddy, and Chandar}]{madsen2022post}
\bibinfo{author}{A.~Madsen}, \bibinfo{author}{S.~Reddy}, \bibinfo{author}{S.~Chandar},
\newblock \bibinfo{title}{Post-hoc interpretability for neural nlp: A survey},
\newblock \bibinfo{journal}{ACM Computing Surveys} \bibinfo{volume}{55} (\bibinfo{year}{2022}) \bibinfo{pages}{1--42}.
\bibitem[{Linardatos et~al.(2020)Linardatos, Papastefanopoulos, and Kotsiantis}]{linardatos2020explainable}
\bibinfo{author}{P.~Linardatos}, \bibinfo{author}{V.~Papastefanopoulos}, \bibinfo{author}{S.~Kotsiantis},
\newblock \bibinfo{title}{Explainable ai: A review of machine learning interpretability methods},
\newblock \bibinfo{journal}{Entropy} \bibinfo{volume}{23} (\bibinfo{year}{2020}) \bibinfo{pages}{18}.
\bibitem[{Cambria et~al.(2023)Cambria, Malandri, Mercorio, Mezzanzanica, and Nobani}]{cambria2023survey}
\bibinfo{author}{E.~Cambria}, \bibinfo{author}{L.~Malandri}, \bibinfo{author}{F.~Mercorio}, \bibinfo{author}{M.~Mezzanzanica}, \bibinfo{author}{N.~Nobani},
\newblock \bibinfo{title}{A survey on xai and natural language explanations},
\newblock \bibinfo{journal}{Information Processing \& Management} \bibinfo{volume}{60} (\bibinfo{year}{2023}) \bibinfo{pages}{103111}.
\bibitem[{Chen et~al.(2023)Chen, Sinavski, H{\"u}nermann, Karnsund, Willmott, Birch, Maund, and Shotton}]{chen2023driving}
\bibinfo{author}{L.~Chen}, \bibinfo{author}{O.~Sinavski}, \bibinfo{author}{J.~H{\"u}nermann}, \bibinfo{author}{A.~Karnsund}, \bibinfo{author}{A.~J. Willmott}, \bibinfo{author}{D.~Birch}, \bibinfo{author}{D.~Maund}, \bibinfo{author}{J.~Shotton},
\newblock \bibinfo{title}{Driving with llms: Fusing object-level vector modality for explainable autonomous driving},
\newblock \bibinfo{journal}{arXiv preprint arXiv:2310.01957}  (\bibinfo{year}{2023}).
\bibitem[{Cui et~al.(2024)Cui, Ma, Cao, Ye, Zhou, Liang, Chen, Lu, Yang, Liao et~al.}]{cui2024survey}
\bibinfo{author}{C.~Cui}, \bibinfo{author}{Y.~Ma}, \bibinfo{author}{X.~Cao}, \bibinfo{author}{W.~Ye}, \bibinfo{author}{Y.~Zhou}, \bibinfo{author}{K.~Liang}, \bibinfo{author}{J.~Chen}, \bibinfo{author}{J.~Lu}, \bibinfo{author}{Z.~Yang}, \bibinfo{author}{K.-D. Liao}, et~al.,
\newblock \bibinfo{title}{A survey on multimodal large language models for autonomous driving},
\newblock in: \bibinfo{booktitle}{Proceedings of the IEEE/CVF Winter Conference on Applications of Computer Vision}, \bibinfo{year}{2024}, pp. \bibinfo{pages}{958--979}.
\bibitem[{Zeiler and Fergus(2014)}]{zeiler2014visualizing}
\bibinfo{author}{M.~D. Zeiler}, \bibinfo{author}{R.~Fergus},
\newblock \bibinfo{title}{Visualizing and understanding convolutional networks},
\newblock in: \bibinfo{booktitle}{Computer Vision--ECCV 2014: 13th European Conference, Zurich, Switzerland, September 6-12, 2014, Proceedings, Part I 13}, \bibinfo{organization}{Springer}, \bibinfo{year}{2014}, pp. \bibinfo{pages}{818--833}.
\bibitem[{Selvaraju et~al.(2017)Selvaraju, Cogswell, Das, Vedantam, Parikh, and Batra}]{selvaraju2017grad}
\bibinfo{author}{R.~R. Selvaraju}, \bibinfo{author}{M.~Cogswell}, \bibinfo{author}{A.~Das}, \bibinfo{author}{R.~Vedantam}, \bibinfo{author}{D.~Parikh}, \bibinfo{author}{D.~Batra},
\newblock \bibinfo{title}{Grad-cam: Visual explanations from deep networks via gradient-based localization},
\newblock in: \bibinfo{booktitle}{Proceedings of the IEEE International Conference on Computer Vision}, \bibinfo{year}{2017}, pp. \bibinfo{pages}{618--626}.
\bibitem[{Bahdanau et~al.(2014)Bahdanau, Cho, and Bengio}]{bahdanau2014neural}
\bibinfo{author}{D.~Bahdanau}, \bibinfo{author}{K.~Cho}, \bibinfo{author}{Y.~Bengio},
\newblock \bibinfo{title}{Neural machine translation by jointly learning to align and translate},
\newblock \bibinfo{journal}{arXiv preprint arXiv:1409.0473}  (\bibinfo{year}{2014}).
\bibitem[{Hadi et~al.(2023)Hadi, Qureshi, Shah, Irfan, Zafar, Shaikh, Akhtar, Wu, Mirjalili et~al.}]{hadi2023survey}
\bibinfo{author}{M.~U. Hadi}, \bibinfo{author}{R.~Qureshi}, \bibinfo{author}{A.~Shah}, \bibinfo{author}{M.~Irfan}, \bibinfo{author}{A.~Zafar}, \bibinfo{author}{M.~B. Shaikh}, \bibinfo{author}{N.~Akhtar}, \bibinfo{author}{J.~Wu}, \bibinfo{author}{S.~Mirjalili}, et~al.,
\newblock \bibinfo{title}{A survey on large language models: Applications, challenges, limitations, and practical usage},
\newblock \bibinfo{journal}{Authorea Preprints}  (\bibinfo{year}{2023}).
\bibitem[{Kroeger et~al.(2023)Kroeger, Ley, Krishna, Agarwal, and Lakkaraju}]{kroeger2023large}
\bibinfo{author}{N.~Kroeger}, \bibinfo{author}{D.~Ley}, \bibinfo{author}{S.~Krishna}, \bibinfo{author}{C.~Agarwal}, \bibinfo{author}{H.~Lakkaraju},
\newblock \bibinfo{title}{Are large language models post hoc explainers?},
\newblock \bibinfo{journal}{arXiv preprint arXiv:2310.05797}  (\bibinfo{year}{2023}).
\bibitem[{Yang et~al.(2024)Yang, Zhu, Wang, Chen, Zhao, Wang, Han, Zheng, and Yin}]{yang2024fine}
\bibinfo{author}{M.~Yang}, \bibinfo{author}{M.~Zhu}, \bibinfo{author}{Y.~Wang}, \bibinfo{author}{L.~Chen}, \bibinfo{author}{Y.~Zhao}, \bibinfo{author}{X.~Wang}, \bibinfo{author}{B.~Han}, \bibinfo{author}{X.~Zheng}, \bibinfo{author}{J.~Yin},
\newblock \bibinfo{title}{Fine-tuning large language model based explainable recommendation with explainable quality reward},
\newblock in: \bibinfo{booktitle}{Proceedings of the AAAI Conference on Artificial Intelligence}, volume~\bibinfo{volume}{38}, \bibinfo{year}{2024}, pp. \bibinfo{pages}{9250--9259}.
\bibitem[{Miller(2019)}]{miller2019explanation}
\bibinfo{author}{T.~Miller},
\newblock \bibinfo{title}{Explanation in artificial intelligence: Insights from the social sciences},
\newblock \bibinfo{journal}{Artificial intelligence} \bibinfo{volume}{267} (\bibinfo{year}{2019}) \bibinfo{pages}{1--38}.
\bibitem[{Hilton(1990)}]{hilton1990conversational}
\bibinfo{author}{D.~J. Hilton},
\newblock \bibinfo{title}{Conversational processes and causal explanation.},
\newblock \bibinfo{journal}{Psychological Bulletin} \bibinfo{volume}{107} (\bibinfo{year}{1990}) \bibinfo{pages}{65}.
\bibitem[{Cawsey(1993)}]{cawsey1993planning}
\bibinfo{author}{A.~Cawsey},
\newblock \bibinfo{title}{Planning interactive explanations},
\newblock \bibinfo{journal}{International Journal of Man-Machine Studies} \bibinfo{volume}{38} (\bibinfo{year}{1993}) \bibinfo{pages}{169--199}.
\bibitem[{Grice(1975)}]{grice1975logic}
\bibinfo{author}{H.~P. Grice},
\newblock \bibinfo{title}{Logic and conversation},
\newblock in: \bibinfo{booktitle}{Speech acts}, \bibinfo{publisher}{Brill}, \bibinfo{year}{1975}, pp. \bibinfo{pages}{41--58}.
\bibitem[{Feine et~al.(2019)Feine, Gnewuch, Morana, and Maedche}]{feine2019taxonomy}
\bibinfo{author}{J.~Feine}, \bibinfo{author}{U.~Gnewuch}, \bibinfo{author}{S.~Morana}, \bibinfo{author}{A.~Maedche},
\newblock \bibinfo{title}{A taxonomy of social cues for conversational agents},
\newblock \bibinfo{journal}{International Journal of Human-Computer Studies} \bibinfo{volume}{132} (\bibinfo{year}{2019}) \bibinfo{pages}{138--161}.
\bibitem[{Apperly and Butterfill(2009)}]{apperly2009humans}
\bibinfo{author}{I.~A. Apperly}, \bibinfo{author}{S.~A. Butterfill},
\newblock \bibinfo{title}{Do humans have two systems to track beliefs and belief-like states?},
\newblock \bibinfo{journal}{Psychological review} \bibinfo{volume}{116} (\bibinfo{year}{2009}) \bibinfo{pages}{953}.
\bibitem[{Hoffman and Klein(2017)}]{hoffman2017explaining}
\bibinfo{author}{R.~R. Hoffman}, \bibinfo{author}{G.~Klein},
\newblock \bibinfo{title}{Explaining explanation, part 1: theoretical foundations},
\newblock \bibinfo{journal}{IEEE Intelligent Systems} \bibinfo{volume}{32} (\bibinfo{year}{2017}) \bibinfo{pages}{68--73}.
\bibitem[{Carvalho et~al.(2019)Carvalho, Pereira, and Cardoso}]{carvalho2019machine}
\bibinfo{author}{D.~V. Carvalho}, \bibinfo{author}{E.~M. Pereira}, \bibinfo{author}{J.~S. Cardoso},
\newblock \bibinfo{title}{Machine learning interpretability: A survey on methods and metrics},
\newblock \bibinfo{journal}{Electronics} \bibinfo{volume}{8} (\bibinfo{year}{2019}) \bibinfo{pages}{832}.
\bibitem[{Hilton(2017)}]{hilton2017social}
\bibinfo{author}{D.~Hilton},
\newblock \bibinfo{title}{Social attribution and explanation},
\newblock \bibinfo{journal}{Oxford Academic}  (\bibinfo{year}{2017}).
\bibitem[{Kahneman and Tversky(1981)}]{kahneman1981simulation}
\bibinfo{author}{D.~Kahneman}, \bibinfo{author}{A.~Tversky}, \bibinfo{title}{The simulation heuristic}, \bibinfo{publisher}{National Technical Information Service}, \bibinfo{year}{1981}.
\bibitem[{Lipton(1990)}]{lipton1990contrastive}
\bibinfo{author}{P.~Lipton},
\newblock \bibinfo{title}{Contrastive explanation},
\newblock \bibinfo{journal}{Royal Institute of Philosophy Supplements} \bibinfo{volume}{27} (\bibinfo{year}{1990}) \bibinfo{pages}{247--266}.
\bibitem[{Molnar(2020)}]{molnar2020interpretable}
\bibinfo{author}{C.~Molnar}, \bibinfo{title}{Interpretable machine learning. 2019}, \bibinfo{year}{2020}.
\bibitem[{Epstude and Roese(2008)}]{epstude2008functional}
\bibinfo{author}{K.~Epstude}, \bibinfo{author}{N.~J. Roese},
\newblock \bibinfo{title}{The functional theory of counterfactual thinking},
\newblock \bibinfo{journal}{Personality and social psychology review} \bibinfo{volume}{12} (\bibinfo{year}{2008}) \bibinfo{pages}{168--192}.
\bibitem[{Anjomshoae et~al.(2021)Anjomshoae, Omeiza, and Jiang}]{anjomshoae2021context}
\bibinfo{author}{S.~Anjomshoae}, \bibinfo{author}{D.~Omeiza}, \bibinfo{author}{L.~Jiang},
\newblock \bibinfo{title}{Context-based image explanations for deep neural networks},
\newblock \bibinfo{journal}{Image and Vision Computing} \bibinfo{volume}{116} (\bibinfo{year}{2021}) \bibinfo{pages}{104310}.
\bibitem[{Ivanovs et~al.(2021)Ivanovs, Kadikis, and Ozols}]{ivanovs2021perturbation}
\bibinfo{author}{M.~Ivanovs}, \bibinfo{author}{R.~Kadikis}, \bibinfo{author}{K.~Ozols},
\newblock \bibinfo{title}{Perturbation-based methods for explaining deep neural networks: A survey},
\newblock \bibinfo{journal}{Pattern Recognition Letters} \bibinfo{volume}{150} (\bibinfo{year}{2021}) \bibinfo{pages}{228--234}.
\bibitem[{Robnik-{\v{S}}ikonja and Bohanec(2018)}]{robnik2018perturbation}
\bibinfo{author}{M.~Robnik-{\v{S}}ikonja}, \bibinfo{author}{M.~Bohanec},
\newblock \bibinfo{title}{Perturbation-based explanations of prediction models},
\newblock \bibinfo{journal}{Human and Machine Learning: Visible, Explainable, Trustworthy and Transparent}  (\bibinfo{year}{2018}) \bibinfo{pages}{159--175}.
\bibitem[{Covert et~al.(2021)Covert, Lundberg, and Lee}]{covert2021explaining}
\bibinfo{author}{I.~Covert}, \bibinfo{author}{S.~Lundberg}, \bibinfo{author}{S.-I. Lee},
\newblock \bibinfo{title}{Explaining by removing: A unified framework for model explanation},
\newblock \bibinfo{journal}{Journal of Machine Learning Research} \bibinfo{volume}{22} (\bibinfo{year}{2021}) \bibinfo{pages}{1--90}.
\bibitem[{Omeiza et~al.(2022)Omeiza, Anjomshoae, Webb, Jirotka, and Kunze}]{omeiza2022spoken}
\bibinfo{author}{D.~Omeiza}, \bibinfo{author}{S.~Anjomshoae}, \bibinfo{author}{H.~Webb}, \bibinfo{author}{M.~Jirotka}, \bibinfo{author}{L.~Kunze},
\newblock \bibinfo{title}{From spoken thoughts to automated driving commentary: Predicting and explaining intelligent vehicles’ actions},
\newblock in: \bibinfo{booktitle}{2022 IEEE Intelligent Vehicles Symposium (IV)}, \bibinfo{organization}{IEEE}, \bibinfo{year}{2022}, pp. \bibinfo{pages}{1040--1047}.
\bibitem[{Anjomshoae et~al.(2021)Anjomshoae, Jiang, and Fr{\"a}mling}]{anjomshoae2021visual}
\bibinfo{author}{S.~Anjomshoae}, \bibinfo{author}{L.~Jiang}, \bibinfo{author}{K.~Fr{\"a}mling},
\newblock \bibinfo{title}{Visual explanations for dnns with contextual importance},
\newblock in: \bibinfo{booktitle}{Explainable and Transparent AI and Multi-Agent Systems: Third International Workshop, EXTRAAMAS 2021, Virtual Event, May 3--7, 2021, Revised Selected Papers 3}, \bibinfo{organization}{Springer}, \bibinfo{year}{2021}, pp. \bibinfo{pages}{83--96}.
\bibitem[{Gao et~al.(2023)Gao, Xiong, Gao, Jia, Pan, Bi, Dai, Sun, and Wang}]{gao2023retrieval}
\bibinfo{author}{Y.~Gao}, \bibinfo{author}{Y.~Xiong}, \bibinfo{author}{X.~Gao}, \bibinfo{author}{K.~Jia}, \bibinfo{author}{J.~Pan}, \bibinfo{author}{Y.~Bi}, \bibinfo{author}{Y.~Dai}, \bibinfo{author}{J.~Sun}, \bibinfo{author}{H.~Wang},
\newblock \bibinfo{title}{Retrieval-augmented generation for large language models: A survey},
\newblock \bibinfo{journal}{arXiv preprint arXiv:2312.10997}  (\bibinfo{year}{2023}).
\bibitem[{Li et~al.(2018)Li, Liu, Xu, and Qiu}]{li2018real}
\bibinfo{author}{B.~Li}, \bibinfo{author}{S.~Liu}, \bibinfo{author}{W.~Xu}, \bibinfo{author}{W.~Qiu},
\newblock \bibinfo{title}{Real-time object detection and semantic segmentation for autonomous driving},
\newblock in: \bibinfo{booktitle}{MIPPR 2017: Automatic Target Recognition and Navigation}, volume \bibinfo{volume}{10608}, \bibinfo{organization}{SPIE}, \bibinfo{year}{2018}, pp. \bibinfo{pages}{167--174}.
\bibitem[{Rodrigues et~al.(2021)Rodrigues, Barbosa, Oliveira~Filho, Cani, Dantas, Sadok, Kelner, Souza, Marquezini, and Lins}]{rodrigues2021modeling}
\bibinfo{author}{I.~R. Rodrigues}, \bibinfo{author}{G.~Barbosa}, \bibinfo{author}{A.~Oliveira~Filho}, \bibinfo{author}{C.~Cani}, \bibinfo{author}{M.~Dantas}, \bibinfo{author}{D.~H. Sadok}, \bibinfo{author}{J.~Kelner}, \bibinfo{author}{R.~S. Souza}, \bibinfo{author}{M.~V. Marquezini}, \bibinfo{author}{S.~Lins},
\newblock \bibinfo{title}{Modeling and assessing an intelligent system for safety in human-robot collaboration using deep and machine learning techniques},
\newblock \bibinfo{journal}{Multimedia Tools and Applications}  (\bibinfo{year}{2021}) \bibinfo{pages}{1--27}.
\bibitem[{Wang et~al.(2017)Wang, Wei, Wang, Gao, Chen, and Shen}]{wang2017hierarchical}
\bibinfo{author}{Z.~Wang}, \bibinfo{author}{L.~Wei}, \bibinfo{author}{L.~Wang}, \bibinfo{author}{Y.~Gao}, \bibinfo{author}{W.~Chen}, \bibinfo{author}{D.~Shen},
\newblock \bibinfo{title}{Hierarchical vertex regression-based segmentation of head and neck ct images for radiotherapy planning},
\newblock \bibinfo{journal}{IEEE Transactions on Image Processing} \bibinfo{volume}{27} (\bibinfo{year}{2017}) \bibinfo{pages}{923--937}.
\bibitem[{Zhou et~al.(2017)Zhou, Lapedriza, Khosla, Oliva, and Torralba}]{zhou2017places}
\bibinfo{author}{B.~Zhou}, \bibinfo{author}{A.~Lapedriza}, \bibinfo{author}{A.~Khosla}, \bibinfo{author}{A.~Oliva}, \bibinfo{author}{A.~Torralba},
\newblock \bibinfo{title}{Places: A 10 million image database for scene recognition},
\newblock \bibinfo{journal}{IEEE Transactions on Pattern Analysis and Machine Intelligence}  (\bibinfo{year}{2017}).
\bibitem[{Chen et~al.(2018)Chen, Zhu, Papandreou, Schroff, and Adam}]{chen2018encoder}
\bibinfo{author}{L.-C. Chen}, \bibinfo{author}{Y.~Zhu}, \bibinfo{author}{G.~Papandreou}, \bibinfo{author}{F.~Schroff}, \bibinfo{author}{H.~Adam},
\newblock \bibinfo{title}{Encoder-decoder with atrous separable convolution for semantic image segmentation},
\newblock in: \bibinfo{booktitle}{Proceedings of the European conference on computer vision (ECCV)}, \bibinfo{year}{2018}, pp. \bibinfo{pages}{801--818}.
\bibitem[{Hao et~al.(2020)Hao, Zhou, and Guo}]{hao2020brief}
\bibinfo{author}{S.~Hao}, \bibinfo{author}{Y.~Zhou}, \bibinfo{author}{Y.~Guo},
\newblock \bibinfo{title}{A brief survey on semantic segmentation with deep learning},
\newblock \bibinfo{journal}{Neurocomputing} \bibinfo{volume}{406} (\bibinfo{year}{2020}) \bibinfo{pages}{302--321}.
\bibitem[{Li et~al.(2018)Li, Shi, Qi, and Chen}]{li2018survey}
\bibinfo{author}{B.~Li}, \bibinfo{author}{Y.~Shi}, \bibinfo{author}{Z.~Qi}, \bibinfo{author}{Z.~Chen},
\newblock \bibinfo{title}{A survey on semantic segmentation},
\newblock in: \bibinfo{booktitle}{2018 IEEE International Conference on Data Mining Workshops (ICDMW)}, \bibinfo{organization}{IEEE}, \bibinfo{year}{2018}, pp. \bibinfo{pages}{1233--1240}.

\end{thebibliography}

\appendix



\end{document}